# GraphBrep: Learning B-Rep in Graph Structure for Efficient CAD Generation

Weilin Lai[1, 2]; Tie Xu[3]; Hu Wang[1, 2, *]

1   *State Key Laboratory of Advanced Design and Manufacturing for Vehicle, Hunan University, Changsha 410082, PR China*

2   *Beijing Institute of Technology Shenzhen Automotive Research Institute, Shenzhen 518000, PR China*

3   *SAIC GM Wuling Automobile Co., Ltd., Liuzhou 545007, PR China.*

## Abstract

Direct B-Rep generation is increasingly important in CAD workflows, eliminating costly modeling sequence data and supporting complex features. A key challenge is modeling joint distribution of the misaligned geometry and topology. Existing methods tend to implicitly embed topology into the geometric features of edges. Although this integration ensures feature alignment, it also causes edge geometry to carry more redundant structural information compared to the original B-Rep, leading to significantly higher computational cost. To reduce redundancy, we propose ***GraphBrep***, a B-Rep generation model that explicitly represents and learns compact topology. Following the original structure of B-Rep, we construct an undirected weighted graph to represent surface topology. A graph diffusion model is employed to learn topology conditioned on surface features, serving as the basis for determining connectivity between primitive surfaces. The explicit representation ensures a compact data structure, effectively reducing computational cost during both training and inference. Experiments on two large-scale unconditional datasets and one category-conditional dataset demonstrate the proposed method significantly reduces training and inference times (up to 31.3% and 56.3% for given datasets, respectively) while maintaining high-quality CAD generation compared with SOTA.

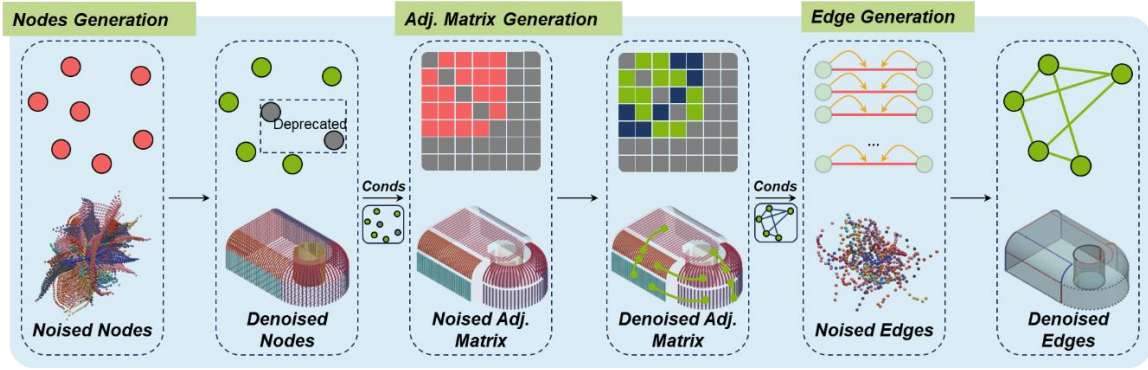

Figure 1 Generation process of our proposed GraphBrep. First, the noisy nodes are denoised to generate surfaces, and duplicate surfaces are removed. Next, conditioned on the nodes, the adjacency matrix, which indicates the topology of surfaces, is generated through a Graph Diffusion model. Finally, based on the adjacency matrix, the information of each pair of endpoint nodes is injected as a conditional input for edge generation.

## 1. Introduction

In modern product design, Computer-Aided Design (CAD) models are indispensable, offering a precise digital link to translate a designer's creativity into tangible outcomes. However, popular CAD modeling remains labor-intensive, particularly when designing intricate geometries. Recent advances in Artificial Intelligence Generated Content (AIGC) have demonstrated promising automation capabilities across various domains, including images[1–6], text[7–10], video[11–13] and 3D object[14–18]. Utilizing these advancements, generative AI models trained on extensive CAD datasets have started automating CAD model generation by effectively learning underlying data distributions[19–26].

A key challenge in generative CAD modeling arises from the complexity inherent in CAD data structures, particularly Boundary Representation (B-Rep). B-Rep defines objects through detailed geometric elements such as surfaces, edges, and vertices, accompanied by complex topological relationships. The structural complexity of B-Rep makes it difficult for traditional neural networks to directly process these data formats. To overcome this, researchers employed simplified representations such as modelling command sequences, converting CAD operations (e.g., "sketch," "extrude," and "fillet") into sequential tokens[24–28]. While this method simplified data structures and makes them more compatible with neural network processing, it significantly limited the range and complexity of generatable features. For instance, the generation of sophisticated surfaces such as Non-Uniform Rational B-Splines (NURBS) remains challenging with such methods. Additionally, modeling command sequences are not standard industry formats, which further complicates the data acquisition process.

Recently, a new paradigm has emerged, focusing on directly processing B-Rep data, which significantly enhanced the generative handling of curves, surfaces, and complex geometries [22,23]. Such methods typically decomposed the complex B-Rep data distribution into multiple simple, independently learned distributions. Topological relationships were implicitly encoded within geometric components, which were later reconstructed according to predefined rules. However, despite these notable advancements, achieving a balance between generational diversity and computational efficiency remains a significant unresolved issue. SolidGen[23], for example, was an early direct B-Rep generative method that tokenized geometric elements but struggled with limited diversity due to the constrained range of geometric features. Conversely, BrepGen[22] addressed arbitrary surfaces and curves more effectively but faced efficiency challenges due to redundant topological representations. They encoded topological relationships within tree structures, which ignored edge-sharing among surfaces, causing redundancy and repeated edges, while assigning surfaces a fixed maximum edge count leads to unnecessary and significant complexity.

To tackle these existing challenges, we propose *GraphBrep*, a novel graph-based diffusion model explicitly designed for efficient and diverse B-Rep CAD generation. *GraphBrep* innovatively represents B-Rep data as weighted undirected graphs in the network, explicitly capturing inherent topological relationships by defining surfaces as nodes and topological connections as edges. This graph-based representation employs adjacency matrices to encode topology efficiently, ensuring the model generates only essential surface connections and significantly reducing redundant edge information.

Specifically, *GraphBrep* learns the B-Rep graph distribution through the following procedure. First, we decompose the complex graph distribution into three sub-distributions: node, adjacency matrix, and edge, similar to SolidGen[23] and BrepGen[22]. The critical difference from previous methods lies in explicitly modeling topology using a graph diffusion model to learn adjacency matrices, replacing the tree-structured topology with redundant edge information by a compact graph representation. Furthermore, conditioned on the learned adjacency matrix, we specifically focus on edges explicitly represented in the adjacency matrix, guiding edge generation by considering the two surfaces sharing each edge. This method avoids lengthy sequences caused by assigning each surface a fixed maximum edge count. Experiments on two large-scale unconditional datasets and one category-conditional dataset demonstrate the proposed method significantly reduces training and inference times (up to 31.3% and 56.3%, respectively) while maintaining high-quality CAD generation compared with state-of-the-art methods.

## 2. Related works

**CAD Command Sequences Generation.** Many studies have focused on developing automated methods for generating CAD modeling command sequences[24–26,29–32]. The central idea behind these approaches is to automate the generation of fundamental modeling commands in CAD models. Commands involve common geometric operations, such as lines, arcs, circles, extrusions and fillets. Each operation is accompanied by specific geometric parameters, such as the coordinates of start and end points of a line, or the center's coordinates and its radius or a circle. The combination of these geometric operations and their corresponding parameters forms a complete command. These commands are then combined in a specific order to construct more geometric structures. Since these command sequences and their associated parameters can be represented as discrete tokens, the CAD model generation problem can be reframed as a task similar to text generation.

By leveraging successful techniques from the natural language processing (NLP) domain, especially the use of Transformer models, researchers have been able to map the generation of CAD command sequence into a text generation framework. Transformer models, known for their ability to handle long-range dependencies and complex sequential structures, excel at generating continuous sequences of CAD commands.

Despite many achievements, CAD command-based generation methods are currently limited to handling simple modeling sequences. These methods face challenges when dealing with complex geometric structures or advanced design operations. Additionally, CAD sequence data is expensive. Most CAD models available in circulation are represented in B-Rep, a general format, due to the gap between different commercial CAD software.

**Direct B-Rep Generation.** Limited work has focused on the direct generation of B-Reps [22,23,33]. Direct B-Rep generation requires handling surface, edge, and vertex features, and assembling these features according to specific relationships, known as geometric features and topological relationships. The geometric features of a B-Rep are represented by parametric expressions, which existing neural networks find difficult to learn directly. While the topological relationships are always dependent on the geometric features.

SolidGen[23] employs transformer-based autoregressive models that first learns the geometry. They first learn distribution of tokenized vertices in the B-Rep. Then, they learn edge distribution conditioned on vertices, which can be intuitively considered as finding the included vertices set and type of each edge. The last is the distribution of surfaces conditioned on vertices and edges, that is, searching for the edges set and type of each surface. By doing this, topological relationships are inherently involved. However, geometric features are still limited here like the methods of command sequences generation.

BrepGen[22] overcomes the limitation of geometric forms. It first discretizes the geometry into discrete point sets and uses diffusion models to learn the surface distribution. Then, conditioned on the surface, it generates all possible geometric edges and vertices for each surface. Finally, these discrete features are stitched back into the B-Rep. Besides, the method learns the topological relationships of the body-surface-edge-vertex tree structure. The discretization allows for any form of geometric features. However, despite its great performance in generation, the tree structure introduces data redundancy, leading to high training and inference costs. Instead of learning the tree-structured topologies, our proposed method directly handles the graph-structured topologies, which reduces data redundancy and lowers the cost.

**Graph Generative Diffusion Models.** The impressive success of diffusion models in image generation has drawn significant interest from researchers in graph generation. Many diffusion or score-based models are carried out for this field. Early work focus on topological relationships. Niu et al. [34] first proposed EDP-GNN, a score-based generative model for graph, which estimates the score function of the adjacency matrix of the graph, and samples the scores at each step by Langevin dynamics. For generate permutation invariant adjacency matrices, Graph-GDP [35] design a Position-enhanced Graph Score Network (PGSN) where the structure and position information of discrete graphs are used for estimated scores on perturbed graphs. Subsequently, more and more researchers attempt to generate node and adjacency matrices simultaneously. Jo et al. [3] proposed GDSS, extending from EDP-GNN, to learn the gradient of the joint log-density for the node and adjacency matrices. Vignac et al. [36] proposed DiGress, which use a graph transformer model as backbone for the denoiser for discrete graph diffusion. NVDiff [37] follows Latent Diffusion [38] and encodes the graphs into latent variables with a VAE model, training a denoiser in a smaller latent space to reduce reduces computation consumptions. To learn the topological relationships in B-Rep, we follow DiGress [36], introduced a Graph Transformer as a denoiser in continuous version for generating the adjacency matrices.

## 3. Methodology

### 3.1 Graph-based B-Rep Representation

In this section, we present the graph-based representation of the B-Rep within GraphBrep. A B-Rep graph is represented as $G = (S, E, A)$, where $S$ denotes surfaces, $E$ represents geometric edges, and $A$ is an adjacency matrix.

For the geometric parts, we follow the approach introduced by UV-Net[39] to align the data that to discretize geometric elements into point sets, which eliminates the inconsistencies between geometric data formats, enabling the neural network to handle different features within a unified model.

The geometric surfaces and edges are discretized into mesh points following UV-Net[39]. The mesh points are sampled in UV parameter domain. The UV domain is a two-dimensional parameter domain, and each geometry in the B-Rep is controlled by a parametric equation in the UV domain, with two independent parameters, U and V range from $[u_{min}, u_{max}] \in \mathbb{R}$ and $[v_{min}, v_{max}] \in \mathbb{R}$.

First, the surface $S$ in the natural domain is mapped to the UV domain, as shown in Figure 2a. First, the shape in the natural domain is transformed into a two-dimensional parametric space in the UV domain. Then, the surface in the UV domain is discretized into an $M \times N$ grid of points. The grid points are uniformly distributed with step sizes $\delta u = \frac{u_{max}-u_{min}}{M-1}$ at $U$ direction and $\delta v = \frac{v_{max}-v_{min}}{N-1}$ at $V$ directions. Finally, each surface is mapped back to the natural domain as point set $p_S$.

Like the surface, the edge is also mapped to the UV domain and discretized into $R$ points with a step size of $\delta u = \frac{u_{max}-u_{min}}{R-1}$ along the U direction. Subsequently, discrete points are mapped back to the natural domain and represented as $p_E$. Vertices are not explicitly considered as separate entities in our approach. This is because vertices naturally are contained as endpoints of edges, serving as the start and end points of each edge.

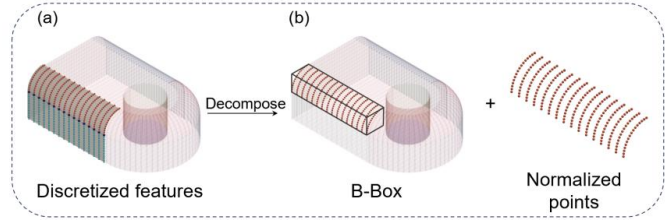

Figure 2 Discretized features of B-Rep. Surfaces and edges are discretized uniformly into $M \times N$ points and $R$ points, respectively, shown in (a). The discretized features are then decomposed to the combination of B-Boxes and normalized points, as shown in (b).

In the following works, $M = N = R = 32$. Additionally, each B-Rep model is translated and scaled into a $[-1,1]^3$ box.

To further reduce the computational complexity of diffusion models and enhance generation efficiency, we followed the approach proposed by BrepGen[22], employing VAE (Variational Autoencoder) to encode the geometric data. For each surface, the geometry is divided into two parts: B-Box $b_S$ and normalized points $b_{S\,normed}$, where $b_S \in \mathbb{R}^{2\times 3}$ represents the diagonal vertices positions of B-Box and $p_{S\,normed}$ is the point set that normalized to $[-1,1]^3$, which is shown in Figure 2(b). The original point set $p_S$ and the normalized point set $p_{S\,normed}$ can be converted back and forth using the B-Box $b_S$. Then, A Surf-VAE is employed to encode the $p_{S\ normed}$, with the latent variable represented as $z_S \in \mathbb{R}^{4\times4\times3}$. Similarly, for each edge, an Edge-VAE encodes the normalized point set $p_{S\ normed}$ into a latent variable $z_E \in \mathbb{R}^{4\times3}$. The surface $s$ and edge $e$ are represented as follows:

$$s = [b_S, z_S] \quad (1)$$

$$e = [b_E, z_E] \quad (2)$$

In a single B-Rep model $G = (S, E, A)$, we maintain three essential components: surface data $S = [s_1, s_2, \ldots, s_{nS}]$, edge data $E = [e_1, e_2, \ldots, e_{nE}]$, and a surface adjacency matrix $A \in \mathbb{R}^{n_S \times n_S}$, where $n_S$ and $n_E$ are the number of surfaces and edges in the model, respectively.

Specifically, we introduce an undirected weighted adjacency matrix $A \in \mathbb{R}^{n_S \times n_S}$ to explicitly demonstrate the pairwise connectivity among all the surfaces, as shown in Figure 3.

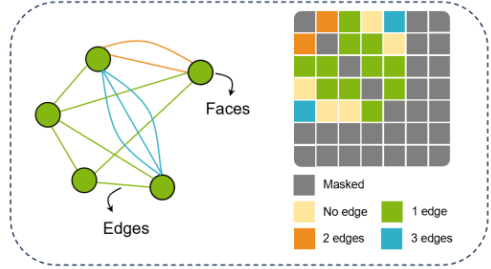

Figure 3 The adjacency matrix in our B-Rep learning. Each element in the matrix demonstrates the number of edges between the corresponding surfaces.

Here we followed SolidGen [23] that split closed faces (such as cylinders) along their seams. Therefore, B-Rep would contain no loops, and the adjacency matrix $A$ is can be simplified as an undirected graph. In this way, $A$ is represented by a $n_S^2$ matrix with $A[i, j] \in [0, e_{max}] \in \mathbb{Z}$. Each element of the matrix indicates the number of shared edges between two surfaces, where $e_{max}$ is the maximum number. Besides, the undefined surfaces are masked, and the diagonal of the matrix is also masked to exclude self-connections.

The other issue is about topology between edges. Unlike the face adjacency matrix, the edge adjacency matrix is not generated through the generative model. The edge topology is divided into two parts: shared vertices within a single surface, and shared vertices between multiple surfaces. For single surface, a geometry-based method is applied. Specifically, the nearest vertices from different edges are regarded as shared vertices. As for multiple surfaces, the shared vertices are be obtained from $A$.

### 3.2 Graph Generative Diffusion Model for B-Rep Generation

In this section, we present our graph generative diffusion model, *GraphBrep*, for efficient B-Rep Generation. Our objective is to learn the distribution $p(G)$ of the B-Rep graph $G(S, E, A)$, where $S$, $E$, and $A$ represent the nodes, edges and the adjacency matrix, respectively. However, unlike pictures or text, directly learning $p(G)$ is challenging due to the complex geometry and the graph data structure. Therefore, we decompose $p(G)$ into serval simple distributions. The model is detailed in the following section.

#### 3.2.1 B-Rep Graph Distribution

The distribution of B-Rep $p(G)$ is first decomposed to the product of three distributions. The unconditional forms are given as follows:

$$p(G) = p(S, E, A) = p(E|A, S) \cdot p(A|S) \cdot p(S) \qquad (3)$$

where $p(S)$ represents the distribution of nodes, $p(A|S)$ is the distribution of the adjacency matrix conditioned on the nodes, and $p(E|A,S)$ represents the distribution of edges conditioned on both the nodes and the adjacency matrix.

Furthermore, $p(S)$ and $p(E)$ are decomposed as follows:

$$p(S) = p(Z_S|B_S) \cdot p(B_S) \tag{4}$$

$$p(E) = p(Z_E|B_E) \cdot p(B_E) \tag{5}$$

Eq. (3) is rewritten as:

$$p(G) = p(Z_E|B_E, A, Z_S, B_S) \cdot p(B_E|A, Z_S, B_S) \cdot p(A|Z_S, B_S) \cdot p(Z_S|B_S) \cdot p(B_S) \tag{6}$$

Decomposing $p(G)$ into several distributions for learning significantly simplifies the distribution learning. Additionally, the learning of each distribution can be performed in parallel, speeding up the overall training process.

### 3.2.2 Diffusion Denoising model

Diffusion model[1] generates data by gradually adding noise and learning a reverse process to restore it. The diffusion model consists of a forward diffusion process and a reverse diffusion process.

**Forward Diffusion Process.** Given a real data point $\mathbf{x_0}$ sample from $q(\mathbf{x})$, Gaussian noise is added at each time step $t \in [1, T]$ to create a sequence of noisy data points $\mathbf{x}_1, \ldots, \mathbf{x}_t$. The noise level at each step is controlled by a variance schedule $\{\beta_t \in (0,1)\}_1^T$, and the forward diffusion process follows the Eq.(7), (8) and (9):

$$q(\mathbf{x}_t|\mathbf{x}_{t-1}) = \mathcal{N}(\mathbf{x}_t; \sqrt{1-\beta_t}\mathbf{x}_t, \beta_t \mathbf{I}) \tag{7}$$

$$q = \mathcal{N}(\mathbf{x}_t; \sqrt{1-\beta_t}\mathbf{x}_t, \beta_t \mathbf{I}) \tag{8}$$

$$q(\mathbf{x}_{1:T}|\mathbf{x}_0) = \prod_{t=1}^{T} q(\mathbf{x}_t|\mathbf{x}_{t-1}) \tag{9}$$

In practice, we sample $\mathbf{x}_t$ by Eq. (10), where $\alpha_t = 1 - \beta_t$, $\bar{\alpha}_t = \prod_{t=1}^{T} \alpha_i$ and $\epsilon \sim \mathcal{N}(0, \mathbf{I})$.

$$\mathbf{x}_t = \sqrt{\bar{\alpha}_t}\mathbf{x}_0 + \sqrt{1 - \bar{\alpha}}\epsilon \tag{10}$$

**Reverse Diffusion Process.** The reverse diffusion process aims to obtain a distribution $p_\theta(\mathbf{x}_{t-1}|\mathbf{x}_t) = \mathcal{N}(\mathbf{x}_{t-1}; \mu_\theta(\mathbf{x}_t, t), \Sigma_\theta(\mathbf{x}_t, t))$ for $t \in [1, T]$ that allows us to sample from the random Gaussian noise. The variance is set to $\Sigma_\theta(\mathbf{x}_t, t) = \sigma_t^2 \mathbf{I}$, and $\sigma_t^2 = \beta_t$ furthermore. Supposed $\mathbf{x}_0$ is known, the reverse distribution can be written as $q(\mathbf{x}_{t-1}|\mathbf{x}_t, \mathbf{x}_0) = \mathcal{N}(\mathbf{x}_{t-1}; \tilde{\mu}(\mathbf{x}_t, \mathbf{x}_0), \tilde{\beta}_t \mathbf{I})$. By using Bayes' theorem, we can obtain $\tilde{\mu}(\mathbf{x}_t, \mathbf{x}_0)$ and $\tilde{\beta}_t$ as Eq. (11) and (12):

$$\tilde{\mu}(\mathbf{x}_t, \mathbf{x}_0) = \frac{\sqrt{\alpha_t}(1 - \bar{\alpha}_{t-1})}{1 - \bar{\alpha}_{t-1}}\mathbf{x}_t + \frac{\sqrt{\bar{\alpha}_{t-1}}\beta_t}{1 - \bar{\alpha}_t} \tag{11}$$

$$\tilde{\beta}_t = \frac{1 - \bar{\alpha}_{t-1}}{1 - \bar{\alpha}_t}\beta_t \tag{12}$$

By rearranging Eq. (11) as $\mathbf{x}_0 = \frac{1}{\sqrt{\bar{\alpha}_t}}(\mathbf{x}_t - \sqrt{1 - \bar{\alpha}}\epsilon)$, we can get the expression of $\mu$:

$$\mu_\theta(\mathbf{x}_t, t) = \frac{1}{\sqrt{\bar{\alpha}_t}}\left(\mathbf{x}_t - \frac{\beta_t}{\sqrt{1 - \bar{\alpha}_t}}\epsilon_\theta(\mathbf{x}_t, t)\right) \tag{13}$$

where $\epsilon_\theta$ is a neural network with trainable parameters $\theta$ to predict $\epsilon$ from $\mathbf{x}_t$, and the loss function $L(\theta) := \mathbb{E}_{t, \mathbf{x}_0, \epsilon}\left[\|\epsilon - \epsilon_\theta(\sqrt{\bar{\alpha}_t}\mathbf{x}_0 + \sqrt{1 - \bar{\alpha}}\epsilon, t)\|^2\right]$. Thus, we can train the model to predict $\epsilon_t$. For generations, we first sample $\mathbf{x}_T \sim \mathcal{N}(0, \mathbf{I})$ and iteratively denoise following Eq. (13).

### 3.2.3 Denoising Networks

*Nodes Denoising.* For the denoising of $p(S)$, we need to learn the distributions $p_\theta(\epsilon^t|B_S{}^t)$ and $p_\theta(\epsilon^t|Z_S{}^t, B_S{}^0)$, where $B_S{}^t$ and $Z_S{}^t$ represent the B-boxes and the latent variables of points of surfaces at timestep $t$, respectively. We followed the DiT [6] approach, employing two Transformer-backbone-based denoisers to model these distributions. The input data is first processed through MLP modules, and then passed into $N$ Transformer encoder layers, which are composed of Multi-head Attention mechanism and Linear layers. The MLP module is used for initial feature extraction, while the Transformer layers, through the self-attention mechanism, capture global dependencies in the data. The conditioned $t$ and $B_S^0$ is embedded in by directly adding after the MLP modules. Finally, these two networks output $B_S^{t-1}$ and $Z_S^{t-1}$, respectively. For simplicity, we will no longer mark $t$ when $t = 0$. Network structure is shown in Figure 4.

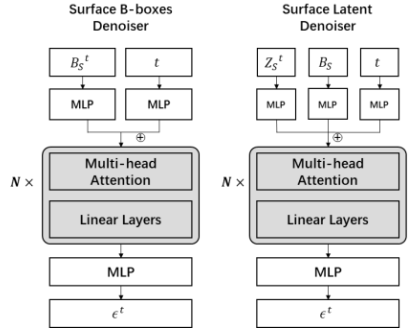

Figure 4 Network structure for nodes denoising.

*Adjacency Matrix Denoising.* We use the noiseless intact nodes as conditions to generate the adjacency matrix. The conditional distribution of the adjacency matrix at time $t$ is expressed as:

$$p_\theta(\epsilon^t|A^t, Z_S, B_S) \qquad (14)$$

Our network for adjacency matrix denoising is Graph Transformer from DiGress[36]. The network structure is shown in Figure 6. However, unlike DiGress, since we learn the graph's nodes, adjacency matrix, and edge attributes in a stepwise manner, we do not output new node features $X_{new}$ and $y_{new}$ at the end. In the Graph Transformer layer, self-attention is first computed on the node features, which can be intuitively understood as the similarity between the nodes. This property is particularly well-suited for generating the adjacency matrix, as the adjacency matrix is essentially derived from the relationships between nodes. Subsequently, cross-attention is applied to update $E$, $X$ and $y$, which represent the latent of adjacency matrix, node features and timestep in the Graph Transformer layer, respectively. The cross-attention mechanism allows the model to effectively integrate information from the different features. After the Graph Transformer layers, the network outputs $E_{new}$. It is important to note that the adjacency matrix is symmetric, so the noise added to the adjacency matrix during the diffusion process must also be symmetric. Additionally, the output $\epsilon^t$ needs to remain symmetric at each timestep. Network structure is shown in Figure 5.

The adjacency matrix denoising network ultimately generates the adjacency matrix $A$, where each element $A[i,j]$ represents the number of shared edges between surface $i$ and surface $j$.

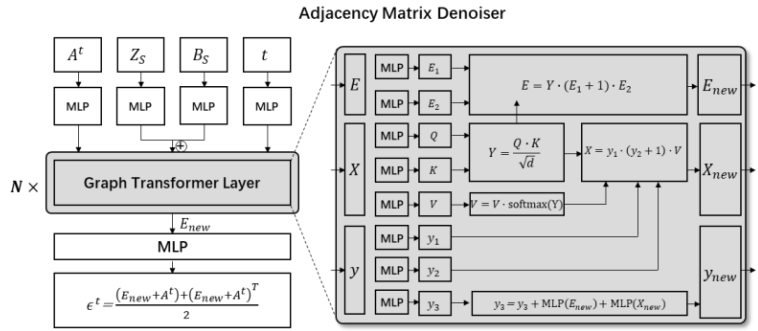

Figure 5 Network structure for Adjacency matrix denoising.

***Edges Denoising.*** Once we have obtained the nodes and the adjacency matrix, we can learn the edge B-boxes distribution $p_\theta(\epsilon^t | B_E{}^t, A, Z_S, B_S)$ and the edge latent distribution $p_\theta(\epsilon^t | Z_E{}^t, B_E, A, Z_S, B_S)$. The edge denoising model uses the same Transformer models as the nodes. First, the surface B-boxes $B_S$ and points latent $Z_S$ of the condition are processed through an MLP module and directly added to $E_B{}^t$ or $E_Z{}^t$ according to the $A$. Since our model represents the B-Rep using an undirected graph, the start and end points of an edge are interchangeable. Therefore, all $B_S$ are be encoded by the same MLP module, and the same applies to the surface $Z_S$.

Here, we use $s_i$ to represent the feature of node $i$, where:

$$s_i = \text{MLP}(b_{S\,i}) + \text{MLP}(z_{S\,i}) + \text{MLP}(t) \qquad (15)$$

For any edge with index $k$ connecting node $i$ and $j$, the features of nodes are added to the edge as Eq. (16) and Eq. (17) for edge B-boxes and latent generation, respectively. In the equations, $b_{Ek}{}^t$ is the noised edge B-box and $z_{Ek}{}^t$ is the noised edge latent indexed $k$ after MLP module.

$$b_{Ek}{}^t = b_{Ek}{}^t + s_i + s_j \qquad (16)$$

$$z_{Ek}{}^t = z_{Ek}{}^t + b_{Ek} + s_i + s_j \qquad (17)$$

Finally, the features are stacked and fed into the Transformer encoder layers, where they are processed through multiple layers and output $\epsilon^t$. The detailed network structure is shown in Figure 6.

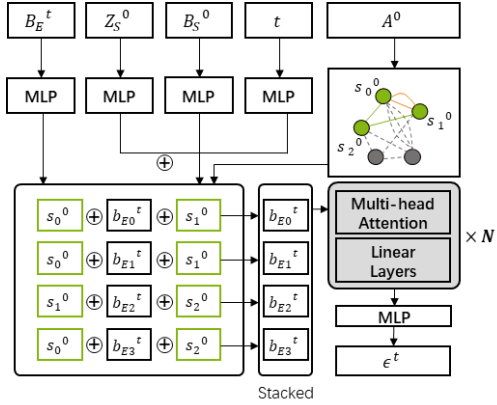

Figure 6 Network structure for edges denoising.

## 4. Experiment

### 4.1 Experiment Setup

#### 4.1.1 Datasets

We use the DeepCAD dataset [26] and A Big CAD Model Dataset For Geometric Deep Learning (ABC) dataset [40] for unconditional generation tasks. The DeepCAD dataset contains approximately 170k CAD models with both modeling sequences and B-Rep representations, ranging from simple solids, such as cubes and cylinders, to more mechanically complex parts like supports and brackets. We follow the hash-based filtering strategy from SolidGen[23] to remove near-duplicate models, duplicated B-Rep models are identified based on the surfaces. Each B-Rep is first centered and normalized. Then, surfaces are discretized into points and quantified into 6-bit. Next, we filtered out B-Rep models with more than 30 faces or with any face having more than 20 edges. Additionally, we restricted the total number of edges to 120. As a result, 85,770 B-Rep models are retained.

The ABC dataset consists of one million models, designed to facilitate research in geometric deep learning. Each model is represented by explicitly parameterized curves and surfaces, offering precise and flexible geometric representations. We select models with less than 50 faces, 30 edges per face, and a total of 150 edges. This process results in a final subset of 258,598 models. Both the data are split into training, validation, and test sets in a 90%-5%-5% ratio.

Additionally, we employ a labeled Furniture dataset[22] containing 10 categories, including chairs, tables, bathtubs, and more. Each model in the Furniture dataset is restricted to a maximum of 50 faces and 150 edges. This dataset allows us to evaluate the conditional generation capabilities of our model, as it provides semantic labels to guide the generation process.

### 4.1.2 Network architecture

**VAE models.** To reduce computational complexity, we first train separate VAE[41] models for surfaces and edges. The surface VAE model consists of 4 down-sampling blocks followed by 4 up-sampling blocks. It compresses the $32 \times 32 \times 3$ discrete surface points into a $4 \times 4 \times 3$ latent space, capturing the essential geometric features while significantly reducing the dimensionality. On the other hand, the edge VAE model has 3 down-sampling blocks and 4 up-sampling blocks. It compresses the $32 \times 3$ discrete edge points into a $4 \times 3$ latent space.

**Node models.** We employ a standard Transformer encoder module with 12 layers and 12 attention heads as the denoiser for the surface B-Boxes. The surface B-Boxes are repeatedly padded to the maximum number of surfaces. The same 12-layer, 12-head Transformer denoiser is applied to the latent. For these two models, the embedded dimension is 1024. In our subsequent comparison tests, we use the same model and checkpoint for both.

**Adjacency matrix model.** We employ a 6-layer, 6-head graph Transformer as the denoiser for the adjacency matrix. The node features, adjacency matrix, and time information are embedded into 768, 64, and 128-dimensional spaces, respectively. To address missing or empty nodes in the matrix, we pad the corresponding entries with zeros, ensuring the data is aligned. Besides, since our adjacency matrix is unidirectional, diagonal elements corresponding to self-relationships of the surfaces were also padded with zeros. The output from the final layer is added to its own transpose, ensuring the predicted noise is symmetric.

**Edge models.** For each edge, both the B-Boxes and the $4 \times 3$ latent of points are zero-padded to the maximum number of edges. We then apply the same denoiser configuration used for surfaces to predict the noise for these edge features.

### 4.1.3 Implementation Detail

In the training part, we implement GraphBrep on 8 Nvidia A100 GPUs (40GB each) using PyTorch as a learning framework. The optimizer used in all models is Adam, with a learning rate of $5e-4$. The loss function is Mean Squared Error (MSE). For the DeepCAD dataset, the batch sizes for surface B-Box, surface latent, adjacency matrix, edge B-Box, and edge latent models are 512, 512, 256, 256, and 256, respectively. The training epochs for these models are 3000, 3000, 2000, 1000, and 1000. For the ABC dataset, the corresponding batch sizes are 512, 256, 256, 256, and 256, with each model trained for 1000 epochs.

For the diffusion process, we use the DDPM scheduler from the Diffusers library [21], with a linear beta schedule ranging from 0.0001 to 0.02, and the model predicts noise. For the five models, the timesteps are set to 1000 for the node and edge models, and 600 for the adjacency matrix model.

The inference process is primarily based on the methodology in [22], where both PNDM and DDPM are used for forward sampling. PNDM provides more efficient sampling compared to DDPM. The initial noise is randomly sampled. For surface B-Box sampling, we first use PNDM for 200 steps of fast sampling and then switch to DDPM for sampling from T=250 to 0. This approach ensures faster and more accurate B-Box sampling. The resulting B-Boxes may contain duplicates, which are removed by comparing the maximum difference in vertex coordinates. If the difference exceeds a threshold of 0.08, the B-Boxes are considered duplicates. After the deduplication, surface latent are generated from the B-Boxes, and 200 steps of PNDM sampling are applied.

For the adjacency matrix sampling, we similarly start with continuous initial noise and perform 250 steps of PNDM sampling. Throughout the sampling process, the adjacency matrix is added to its transpose to ensure symmetry. Furthermore, since each value in the adjacency matrix is a non-negative integer, the final matrix is clipped and rounded to integers. Finally, the dense adjacency matrix is converted into a sparse representation. For edge B-Box and edge latent sampling, PNDM sampling is performed for 250 and 200 steps, respectively.

In the inference part, the maximum sampling faces and edges are 30, 80, 80 and 150, 200, 200 for DeepCAD, ABC and furniture datasets, respectively.

#### 4.1.4 Quantitative Evaluation Metrics

To maintain consistency and comparability with prior work, we adopt the evaluation metrics introduced by previous works[22,23]. We employ two categories of metrics: Distribution Metrics and CAD Metrics.

**Distribution Metrics:**

We randomly sample 3,000 B-Rep models from the generated set and 1,000 B-Rep models from the reference test set. For the comparison of metrics such as coverage and minimum matching distance, we follow previous work and use point cloud representations. Specifically, we randomly sample 2,000 points on the surfaces of the B-Reps.

**(1) Coverage (COV)**, percentage of reference models that have at least one corresponding generated model defined by Chamfer Distance (CD). COV measures how well the generated models cover the diversity of the reference set, ensuring that the generative model can replicate a wide range of shapes present in the test data.

**(2) Minimum Matching Distance (MMD)**, average Chamfer Distance between each reference model and its nearest generated counterpart. MMD quantifies the closeness of the generated models to the reference set, indicating the accuracy of the generative model in replicating specific shapes.

**(3) Jensen-Shannon Divergence (JSD)**, distributional distance between reference and generated data after converting point clouds into $28^3$ discrete voxels. JSD assesses the similarity between the overall distributions of the generated and reference datasets, providing a holistic measure of generative quality beyond individual model comparisons.

**CAD Metrics:**

The same 3000 B-Reps are used to calculate CAD Metrics. The CAD metrics are calculated by the surfaces, which are discretized into $32 \times 32 \times 3$ points and quantized using 6-bit precision.

**(1) Novelty**, percentage of generated B-Reps that do not appear in the training set, indicating the originality of the generated models.

**(2) Uniqueness**, fraction of generated B-Reps that are unique (appear only once in the generation process), reflecting the diversity of the generated set.

**(3) Validity**: percentage of generated B-Reps that form watertight solids, ensuring the geometric correctness and integrity of the models. The encoded B-Rep is synthesized and then validated using the Autodesk Fusion 360 Validation API to ensure its correctness. Models containing intersecting or overlapping faces, as well as invalid faces, are discarded.

## 4.2 Generation

### 4.2.1 Quantitative Evaluation

We select BrepGen [2] and DeepCAD [6] for comparison. To ensure robustness, the tests are repeated 20 times, and the average values were computed. The results are presented in Table 1.

In the comparison of distribution metrics, DeepCAD slightly outperforms our method and BrepGen in terms of COV and JSD. In the MMD comparison, our method performs better, demonstrating stronger generative capability and diversity. Regarding CAD metrics, our method achieves similar performance to BrepGen in terms of Novel and Unique metrics. Additionally, when evaluating Validity that measures whether the generated results can be reconstructed into complete, watertight bodies, our method shows notable improvements, demonstrating the effectiveness of *GraphBrep* in generating valid and feasible B-Rep models.

As for the ABC dataset, since it does not provide modeling sequences and DeepCAD does not support direct processing of STEP files, we only compared our method against BrepGen. In this scenario, our approach demonstrates a slight advantage over BrepGen in terms of COV, MMD, and JSD. Furthermore, our method shows improvements in the feasibility of the generated models.

Table 1 Distribution metrics and CAD metrics for DeepCAD, BrepGen and GraphBrep models on COV, MMD, JSD, Novel, Unique and Valid. DeepCAD dataset and ABC datasets are tested. Both MMD and JSD are multiplied by $10^2$.

| Method | Dataset | COV % ↑ | MMD ↓ | JSD ↓ | Novel % ↑ | Unique % ↑ | Valid % ↑ |
|---|---|---|---|---|---|---|---|
| DeepCAD |  | 74.15 | 1.02 | 1.08 | 87.4 | 89.3 | 66.31 |
| BrepGen | DeepCAD | 73.82 | 0.96 | 1.18 | 99.7 | 99.7 | 63.13 |
| GraphBrep |  | 74.06 | 0.93 | 1.20 | 99.6 | 99.8 | 66.62 |
| BrepGen | ABC | 62.50 | 1.35 | 2.39 | 99.7 | 99.4 | 53.56 |
| GraphBrep |  | 62.55 | 1.34 | 2.35 | 99.7 | 99.6 | 55.72 |

### 4.2.2 Qualitative Evaluation

Figure 7 shows the sampling results of DeepCAD, BrepGen, and our method. By analyzing these results, we observe that our method and BrepGen, the direct B-Rep generation methods, are capable of generating more complex geometries compared to DeepCAD, which is based on modeling command sequences.

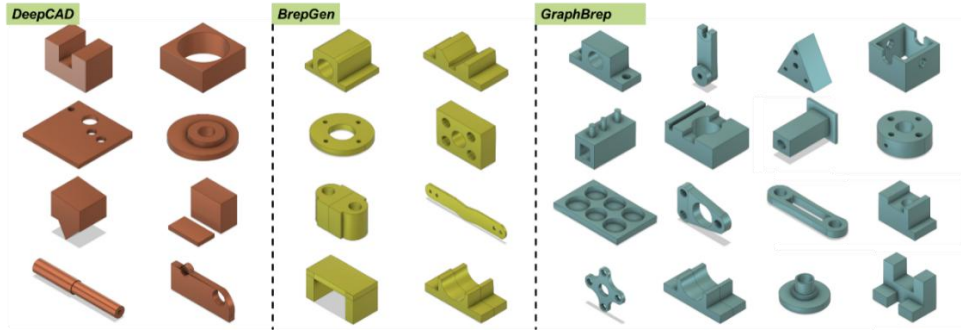

Figure 7 Sampling results of DeepCAD, BrepGen and GraphBrep under DeepCAD datasets.

We further compare the generated results of different complexities, using the number of surfaces as a metric for complexity. Figure 8 shows the distribution of complexity for the sampled results. It is evident that the complexity distribution of all methods closely aligns with the training set. Due to the prevalence of rectangular cuboids in the dataset, the B-Rep models with six surfaces have the highest proportion. The sampling results with $0-15$ faces and $15-30$ faces are shown in Figure 9. We observe that direct B-Rep generation methods produce results with more geometric plausibility. In the sampling results of the DeepCAD model, solids often overlap. Although these models are processed by Open CASCADE to ensure watertightness, this does not guarantee their geometric validity or plausibility.

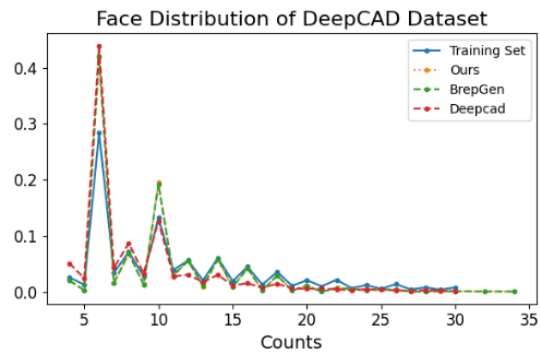

Figure 8 Distribution of complexity of sampled results.

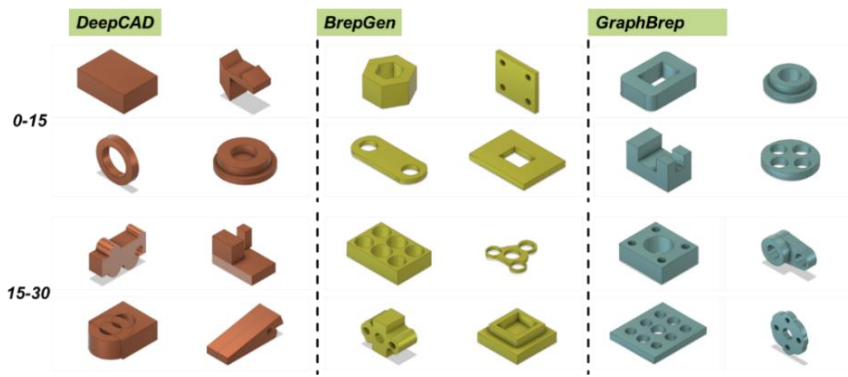

Figure 9 Sampling results of DeepCAD, BrepGen and GraphBrep under DeepCAD datasets in different face amounts.

We also compare the results on the ABC dataset, as shown in Figure 10. The distributions in both datasets are quite similar. This is because both methods share the same surface models when sampling surfaces. We can see that both methods are capable of generating B-Rep models with complex structures.

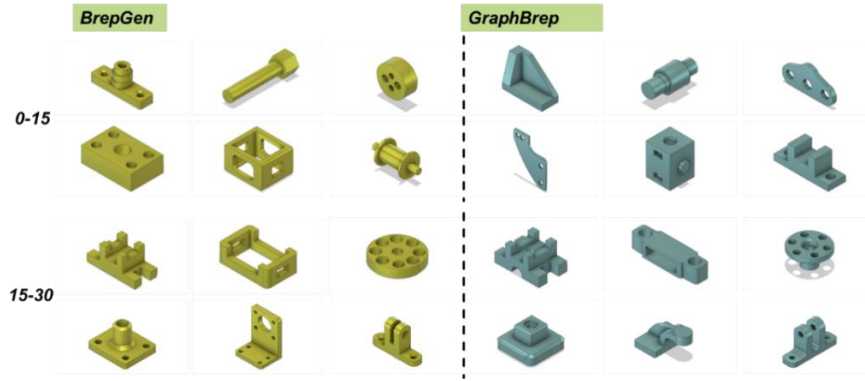

Figure 10 Sampling results of BrepGen and GraphBrep under ABC datasets in different face amounts.

### 4.2.3 Efficiency analysis

One primary distinction between our work and BrepGen [1] lies in the handling of topological relationships. In Section 4.1.2, we discussed how GraphBrep represents the topological relationships between surfaces in a B-Rep using a weighted undirected graph. After generating the nodes (surfaces), we directly use an Adjacency model to generate the adjacency matrix, which then guides the generation of the geometric edges. In contrast, BrepGen [1] assigns a fixed number of edges to each surface after it is generated. The topological relationships between surfaces, edges, and vertices are then established using a matching algorithm, with the core idea being that the distance between two elements determines their relationship. In this approach, the number of edges depends on the surface in the dataset with the maximum number of edges.

For example, in the DeepCAD dataset, during training, the edge number per surface is set to 20, and the number of surfaces is 30. As a result, in the edge B-Boxes and latent models, the sequence length is 600 during training and 900 during inference. In the ABC dataset, the training sequence length is 1500, and the inference sequence length is 3200. Given that the attention mechanism used by BrepGen[22] has a complexity that scales quadratically with the sequence length, this leads to significantly longer training and inference times.

On the other hand, while our method requires the additional training of an adjacency matrix model, once trained, we only need to handle edge sequences of lengths 120 and 150 for the DeepCAD and ABC datasets, respectively. This significantly reduces the complexity of both training and inference. As a result, our method offers a notable efficiency advantage. By optimizing the graph-based representation and leveraging more efficient sampling strategies, we are able to maintain competitive performance while achieving faster execution times with similar parameter settings.

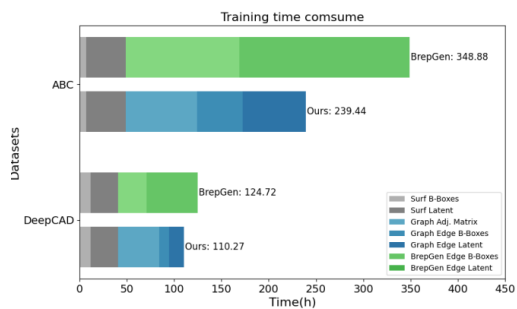

Figure 11 Training time comparison between our method and BrepGen on the DeepCAD and ABC datasets.

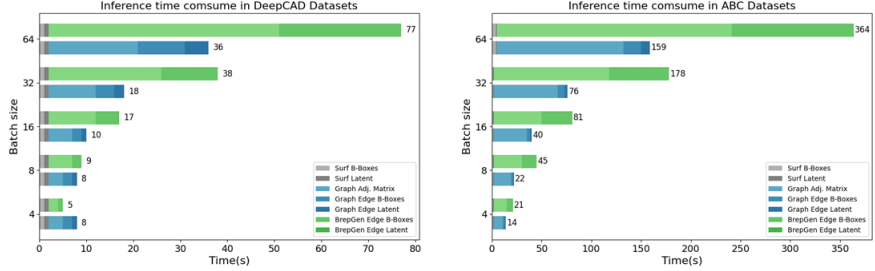

Figure 12 Inference time comparison between our method and BrepGen on the DeepCAD (left) and ABC (right) datasets with varying batch sizes.

In this part, we used one Nvidia RTX 4090 (24G) for an efficiency test. The comparison of training and inference times is presented in Figure 11 and Figure 12, respectively. During training, both methods share the same Surface B-Boxes and Latent models, resulting in identical training times. However, as our method reduces the edge sequence length, the training time for our edge generation model is significantly shorter. Overall, our model demonstrates faster training times across both datasets. The ABC dataset, with its larger number of surfaces and edges, naturally requires more time than the DeepCAD dataset.

Similarly, during the inference process, the inference times for the Surface B-Boxes and Latent models are the same and account for a negligible portion of the total time. This is because the surface sequence is relatively short, leading to faster self-attention computations. For edge inference, our method is generally faster, and this advantage becomes more pronounced as the batch size increases. The difference is especially significant on the ABC dataset, where the edge sequence is longer. The improved inference efficiency with larger batch sizes suggests that our method can provide faster choices for designers and reduce computational requirements, which is particularly beneficial for large-scale tasks.

### 4.2.4 Conditional Generation

We conduct experiments on conditional generation in GraphBrep using classifier-free guidance with Furniture datasets. In this setup, the labels were encoded as vectors and directly added to the inputs of the five networks mentioned above. Figure 13 presents the sampling results under different labels. The results demonstrate that GraphBrep is capable of producing complex geometries while preserving correct topological structures in the conditional setting. Notably, the generated shapes remain diverse and structurally valid across different conditions, indicating the model's adaptability.

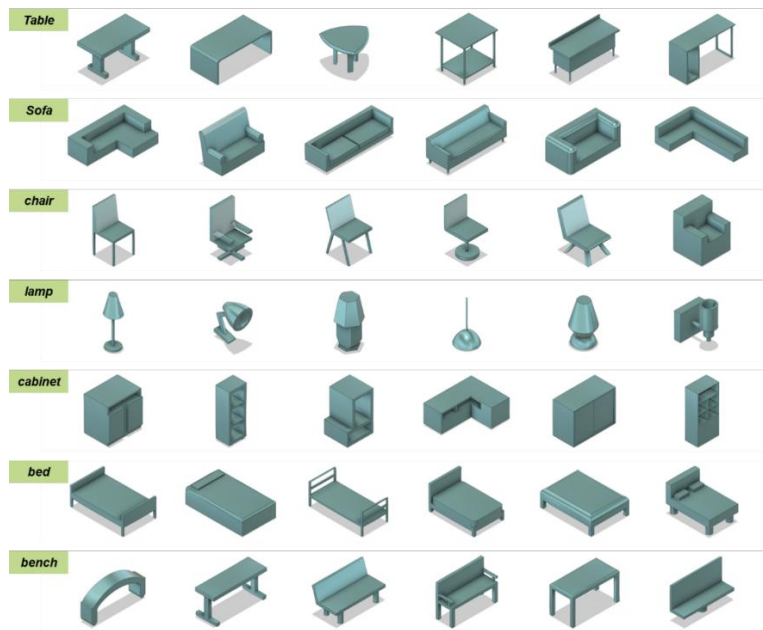

Figure 13 Conditioned sampling results of *GraphBrep* under Furniture datasets.

## 4.3 Ablation Studies

We conduct additional experiments on GraphBrep by varying the maximum allowed surface numbers in generated models. Specifically, we test surface-number limits of 50 and 80 on the ABC dataset, and limits of 50 and 80 on the ABC dataset, and 50 and 80 on the ABC dataset. Quantitative results from these experiments are presented in Table 2. As shown by the quantitative results in Table 3, we observe that as the maximum allowed number of surfaces increases, the quantitative metrics tend to decrease. However, the Novel and Unique metrics remain nearly unchanged. Meanwhile, the validity rate (valid rate) increases.

Table 2 Evaluation of GraphBrep with different surface numbers in sampling.

| Surface Num | Dataset | COV % ↑ | MMD ↓ | JSD ↓ | Novel % ↑ | Unique % ↑ | Valid % ↑ |
|---|---|---|---|---|---|---|---|
| 30 | DeepCAD | 74.06 | 0.93 | 1.20 | 99.6 | 99.8 | 66.62 |
| 50 | | 66.47 | 1.06 | 1.28 | 99.7 | 99.8 | 74.10 |
| 50 | ABC | 68.90 | 1.22 | 1.61 | 99.7 | 99.7 | 45.03 |
| 80 | | 62.55 | 1.34 | 2.35 | 99.7 | 99.6 | 55.72 |

To further clarify the differences, we compare the distribution of surface counts between the generated samples and the original datasets, see Figure 14. From the analysis, we observe in both datasets that increasing the maximum allowed number of surfaces causes a noticeable deviation in quantitative metrics. Specifically, as the preset maximum surface number increases, the sampled distribution shifts toward higher complexity, making the generated samples less aligned with the original dataset, which primarily consists of simpler rectangular geometries. This discrepancy results in a larger deviation in the quantitative metrics. Additionally, we find that when sampling more complex models, the repeated padding strategy mentioned in Section 5.1.2 preserves complete geometric features more effectively, thereby increasing the valid rate for models with higher surface counts.

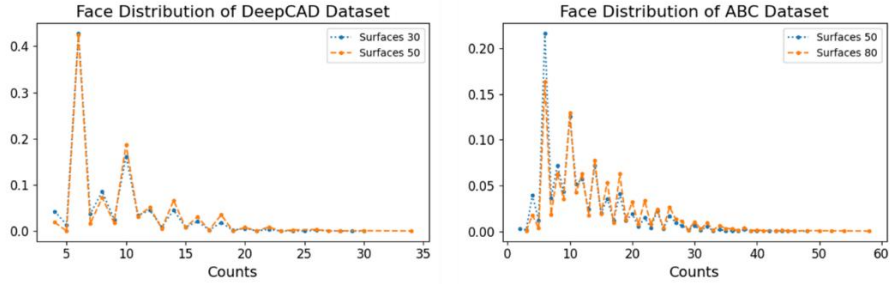

Figure 14 Face distribution under DeepCAD with 30 and 50 surfaces, ABC with 50 and 80 surfaces.

We also test the original DiGress architecture, which simultaneously generates graph nodes and adjacency matrices, aiming to further reduce computational costs. Since this model cannot directly process the complete graph structure of a B-Rep model, we represent surface B-Boxes as graph nodes while keeping the definition of the adjacency matrix unchanged. Regarding the loss function, the losses from both nodes and adjacency matrices were summed directly. We train the model for 4000 epochs. On the DeepCAD dataset, the valid rate does not exceed 15%. Based on our observations, the network struggles to generate correct surface B-Boxes and their corresponding topological relationships. In GraphBrep, the adjacency matrix is generated based on complete surface information, whereas in this setting, only B-Boxes are provided. We believe that this could be a major limitation, as B-Boxes offer only coarse boundary information and fail to capture the precise geometric features of surfaces, thereby hindering the network's ability to learn effective topological structures.

## 5. Conclusion and Limitations

In this paper, we propose GraphBrep, a graph generative diffusion model designed for B-Rep CAD model generation. By representing B-Rep structures as weighted undirected graphs and decomposing the complex joint distribution into conditional distributions over nodes, adjacency matrices, and edges, our method effectively captures both geometric and topological information. Through this design, GraphBrep significantly reduces redundant calculations and improves both training and inference efficiency. Experimental results on DeepCAD, ABC, and Furniture datasets demonstrate that our model not only achieves comparable or superior generation quality relative to state-of-the-art methods but also offers notable computational advantages. Besides, GraphBrep improves the feasibility and validity of generated models while enabling efficient sampling, which is crucial for practical industrial applications. Moreover, our conditional generation experiments show the model's flexibility and robustness in generating diverse and semantically aligned CAD models under different conditions.

While GraphBrep improves efficiency in direct B-Rep generation, several limitations remain. First, the decomposition into multiple sub-models increases system complexity and may cause error accumulation across stages, limiting compactness and robustness. Second, the current model only supports single B-Rep solids and cannot handle assemblies, which are common in real-world CAD tasks. Third, existing evaluation metrics borrowed from point cloud generation fail to reflect the quality and distribution of B-Rep models accurately. In practice, direct B-Rep generation often produces better results than sequence-based methods, but this advantage is not captured numerically, highlighting the need for better evaluation standards.

Gontijo-Lopes, J. Gordon, M. Grafstein, S. Gray, R. Greene, J. Gross, S.S. Gu, Y. Guo, C. Hallacy, J. Han, J. Harris, Y. He, M. Heaton, J. Heidecke, C. Hesse, A. Hickey, W. Hickey, P. Hoeschele, B. Houghton, K. Hsu, S. Hu, X. Hu, J. Huizinga, S. Jain, S. Jain, J. Jang, A. Jiang, R. Jiang, H. Jin, D. Jin, S. Jomoto, B. Jonn, H. Jun, T. Kaftan, Ł. Kaiser, A. Kamali, I. Kanitscheider, N.S. Keskar, T. Khan, L. Kilpatrick, J.W. Kim, C. Kim, Y. Kim, J.H. Kirchner, J. Kiros, M. Knight, D. Kokotajlo, Ł. Kondraciuk, A. Kondrich, A. Konstantinidis, K. Kosic, G. Krueger, V. Kuo, M. Lampe, I. Lan, T. Lee, J. Leike, J. Leung, D. Levy, C.M. Li, R. Lim, M. Lin, S. Lin, M. Litwin, T. Lopez, R. Lowe, P. Lue, A. Makanju, K. Malfacini, S. Manning, T. Markov, Y. Markovski, B. Martin, K. Mayer, A. Mayne, B. McGrew, S.M. McKinney, C. McLeavey, P. McMillan, J. McNeil, D. Medina, A. Mehta, J. Menick, L. Metz, A. Mishchenko, P. Mishkin, V. Monaco, E. Morikawa, D. Mossing, T. Mu, M. Murati, O. Murk, D. Mély, A. Nair, R. Nakano, R. Nayak, A. Neelakantan, R. Ngo, H. Noh, L. Ouyang, C. O'Keefe, J. Pachocki, A. Paino, J. Palermo, A. Pantuliano, G. Parascandolo, J. Parish, E. Parparita, A. Passos, M. Pavlov, A. Peng, A. Perelman, F. de A.B. Peres, M. Petrov, H.P. de O. Pinto, Michael, Pokorny, M. Pokrass, V.H. Pong, T. Powell, A. Power, B. Power, E. Proehl, R. Puri, A. Radford, J. Rae, A. Ramesh, C. Raymond, F. Real, K. Rimbach, C. Ross, B. Rotsted, H. Roussez, N. Ryder, M. Saltarelli, T. Sanders, S. Santurkar, G. Sastry, H. Schmidt, D. Schnurr, J. Schulman, D. Selsam, K. Sheppard, T. Sherbakov, J. Shieh, S. Shoker, P. Shyam, S. Sidor, E. Sigler, M. Simens, J. Sitkin, K. Slama, I. Sohl, B. Sokolowsky, Y. Song, N. Staudacher, F.P. Such, N. Summers, I. Sutskever, J. Tang, N. Tezak, M.B. Thompson, P. Tillet, A. Tootoonchian, E. Tseng, P. Tuggle, N. Turley, J. Tworek, J.F.C. Uribe, A. Vallone, A. Vijayvergiya, C. Voss, C. Wainwright, J.J. Wang, A. Wang, B. Wang, J. Ward, J. Wei, C.J. Weinmann, A. Welihinda, P. Welinder, J. Weng, L. Weng, M. Wiethoff, D. Willner, C. Winter, S. Wolrich, H. Wong, L. Workman, S. Wu, J. Wu, M. Wu, K. Xiao, T. Xu, S. Yoo, K. Yu, Q. Yuan, W. Zaremba, R. Zellers, C. Zhang, M. Zhang, S. Zhao, T. Zheng, J. Zhuang, W. Zhuk, B. Zoph, GPT-4 Technical Report, (2024). https://doi.org/10.48550/arXiv.2303.08774.

[9]  DeepSeek-AI, D. Guo, D. Yang, H. Zhang, J. Song, R. Zhang, R. Xu, Q. Zhu, S. Ma, P. Wang, X. Bi, X. Zhang, X. Yu, Y. Wu, Z.F. Wu, Z. Gou, Z. Shao, Z. Li, Z. Gao, A. Liu, B. Xue, B. Wang, B. Wu, B. Feng, C. Lu, C. Zhao, C. Deng, C. Zhang, C. Ruan, D. Dai, D. Chen, D. Ji, E. Li, F. Lin, F. Dai, F. Luo, G. Hao,
33

39